\documentclass[nohyperref]{article}

\usepackage{microtype}
\usepackage{graphicx}
\usepackage{subfigure}
\usepackage{booktabs} %
\usepackage{enumitem}

\usepackage{hyperref}

\usepackage[accepted]{icml2022}

\usepackage{amsmath}
\usepackage{amssymb}
\usepackage{mathtools}
\usepackage{amsthm}
\usepackage[capitalize,noabbrev]{cleveref}

\theoremstyle{plain}

\theoremstyle{definition}

\theoremstyle{remark}

\DeclareMathOperator*{\argmax}{arg\,max}
\usepackage[textsize=tiny]{todonotes}

\icmltitlerunning{A State-Distribution Matching Approach to Non-Episodic Reinforcement Learning}

\begin{document}

\twocolumn[
\icmltitle{A State-Distribution Matching Approach to Non-Episodic Reinforcement Learning}

\icmlsetsymbol{equal}{*}

\begin{icmlauthorlist}
\icmlauthor{Archit Sharma}{equal,stan}
\icmlauthor{Rehaan Ahmad}{equal,stan}
\icmlauthor{Chelsea Finn}{stan}
\end{icmlauthorlist}

\icmlaffiliation{stan}{Stanford University, CA, USA}

\icmlcorrespondingauthor{Archit Sharma}{architsh@stanford.edu}
\icmlcorrespondingauthor{Rehaan Ahmad}{rehaan@stanford.edu}

\icmlkeywords{reinforcement learning, autonomous, discriminator}

\vskip 0.3in
]

\printAffiliationsAndNotice{\icmlEqualContribution} %

\begin{abstract}
While reinforcement learning (RL) provides a framework for learning through trial and error, translating RL algorithms into the real world has remained challenging. A major hurdle to real-world application arises from the development of algorithms in an episodic setting where the environment is reset after every trial, in contrast with the continual and non-episodic nature of the real-world encountered by embodied agents such as humans and robots.
Enabling agents to learn behaviors autonomously in such non-episodic environments requires that the agent to be able to conduct its own trials.
Prior works have considered an alternating approach where a forward policy learns to solve the task and the backward policy learns to reset the environment, but what initial state distribution should the backward policy reset the agent to? Assuming access to a few demonstrations, we propose a new method, MEDAL, that trains the backward policy to match the state distribution in the provided demonstrations. This keeps the agent close to the task-relevant states, allowing for a mix of easy and difficult starting states for the forward policy. Our experiments show that MEDAL matches or outperforms prior methods on three sparse-reward continuous control tasks from the EARL benchmark, with 40\% gains on the hardest task, while making fewer assumptions than prior works. Code and videos are at: \hyperlink{https://sites.google.com/view/medal-arl/home}{https://sites.google.com/view/medal-arl/home}
\end{abstract}

\section{Introduction}

A cornerstone of human and animal intelligence is the ability to learn autonomously through trial and error. To that extent, reinforcement learning (RL) presents a natural framework to develop learning algorithms for embodied agents. Unfortunately, the predominant emphasis on episodic learning represents a departure from the continual non-episodic nature of the real-world, which presents multiple technical challenges. First, episodic training undermines the autonomy of the learning agent by requiring repeated extrinsic interventions to reset the environment after every trial, which can be both time-consuming and expensive as these interventions may have to be conducted by a human. Second, episodic training from narrow initial state distributions can lead to less robust policies that are reliant on environment resets to recover; e.g. \citet{sharma2021autonomous} show that policies learned in episodic settings with narrow initial state distributions are more sensitive to perturbations than those trained in non-episodic settings.

\begin{figure}[t]
    \centering
    \includegraphics[width=1.0\columnwidth]{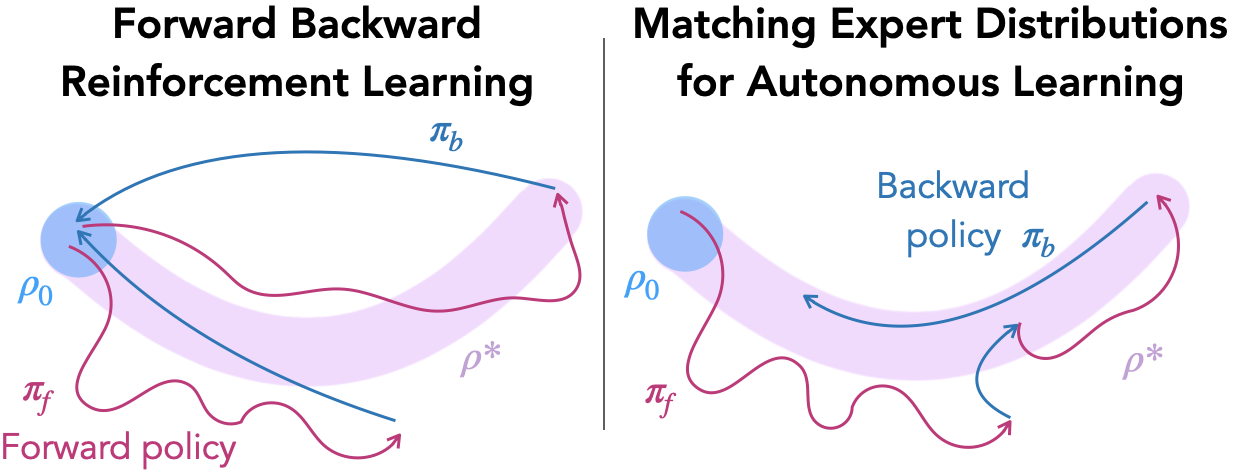}
    \caption{An overview of our proposed method MEDAL (\textit{right}) contrasting it with forward-backward RL~\citep{han2015learning,eysenbach2017leave} (\textit{left}). MEDAL trains a backward policy $\pi_b$ to pull the agent back to the state distribution defined by the demonstrations, enabling the forward policy $\pi_f$ to the learn the task efficiently in contrast to FBRL that retrieves the agent to the initial state distribution before every trial of $\pi_f$.}
    \label{fig:overview}
\end{figure}

Prior works have found that conventional RL algorithms substantially depreciate in performance when applied in non-episodic settings \citep{co2020ecological, zhu20ingredients, sharma2021autonomous}.
Why do such algorithms struggle to learn in non-episodic, autonomous RL (ARL) settings?
Resetting the environment after every single episode allows for natural repetition: the agent can repeatedly practice the task under a narrow set of initial conditions to incrementally improve the policy. Critically, algorithms developed for episodic learning do not have to learn how to reach these initial conditions in the first place. Thus, the main additional challenge in non-episodic, autonomous RL settings is to enable the repetitive practice that is necessary to learn an adept policy. For example, an autonomous robot that is practicing how to close a door will also need to learn how to open a door.

Several recent works learn a backward policy to enable the main forward policy to practice the task: for example, \citet{han2015learning, eysenbach2017leave} propose a backward policy that learns to match the initial state distribution.
However, unlike the episodic setting, the agent can practice the task from any initial state, and not just the narrow initial state distribution that is usually provided by resets. Can the backward policy create starting conditions that enable the forward policy to improve efficiently? It could be useful for the agent to try the task both from ``easy'' states that are close to the goal and harder states that are representative of the starting conditions at evaluation. Easier and harder initial conditions can be seen as a curriculum that simplifies exploration.
\citet{kakade2002approximately} provide a theoretical discussion on how the initial state distribution affects the performance of the learned policy. One of the results show that the closer the starting state distribution is to the \textit{state distribution of the optimal policy} $\rho^*$, the faster the policy moves toward the optimal policy $\pi^*$.
While an oracle access to $\rho^*$ is rarely available, we often have access to a modest set of demonstrations. In this work, we aim to improve autonomous RL by learning a backward policy that matches the starting state distribution to the state distribution observed in the demonstrations.
This enables the agent to practice the task from a variety of initial states, including some that are possibly easier to explore from. An intuitive representation of the algorithm is shown in Figure~\ref{fig:overview}.

The primary contribution of our work is an autonomous RL algorithm \textit{Matching Expert Distributions for Autonomous Learning} (MEDAL), which learns a backward policy that matches the state distribution of a small set of demonstrations, in conjunction with a forward policy that optimizes the task reward. We use a classification based approach that implicitly minimizes the distance between the state distribution of the backward policy and the state distribution in the demonstrations without requiring the density under either distribution.
In Section~\ref{sec:experiments}, we empirically analyze the performance of MEDAL on the Environments for Autonomous RL (EARL) benchmark \citep{sharma2021autonomous}. We find that MEDAL matches or outperforms competitive baselines in all of the sparse-reward environments, with a more than a 40\% gain in success rate on the hardest task where all other comparisons fail completely. Our ablations additionally indicate the importance of matching the state distribution in the demonstrations, providing additional empirical support for the hypothesis that the expert state distribution constitutes a good starting state distribution for learning a task.

\section{Related Work}
\textbf{Autonomous RL.} Using additional policies to enable autonomous learning goes back to the works of \citep{rivest1993inference} in context of finite state automaton, also referred to as ``homing strategies'' in \citep{even2005reinforcement} in context of POMDPs. More recently, in context of continuous control, several works propose autonomous RL methods targeting different starting distributions to learn from: \citet{han2015learning, eysenbach2017leave} match the initial state distribution, \citet{zhu20ingredients} leverage state-novelty \citep{burda2018exploration} to create new starting conditions for every trial, and \citet{sharma2021autonomouscurr} create a curriculum of starting states based on the performance of the forward policy to accelerate the learning. In addition, \citep{xu2020continual, lu2020reset} leverage ideas from unsupervised skill discovery \citep{gregor2016variational, eysenbach2018diversity,sharma2019dynamics, hazan2019provably, campos2020explore},
with the former using it to create an adversarial initial state distribution and the latter to tackle non-episodic lifelong learning with a non-stationary task-distribution. Our work proposes a novel algorithm MEDAL that, unlike these prior works,
opts to match the starting distribution to the state distribution
of demonstrations.
Value-accelerated Persistent RL (VaPRL) \citep{sharma2021autonomouscurr} also considers the problem of autonomous RL with a few initial demonstrations.
Unlike VaPRL, our algorithm does not rely on relabeling transitions with new goals \citep{andrychowicz2017hindsight}, and thus does not require access to the functional form of the reward function, eliminating the need for additional hyperparameters that require task-specific tuning.
A simple and task-agnostic ARL method would accelerate the development of autonomous robotic systems, the benefits of such autonomy being demonstrated by several recent works
\citep{chatzilygeroudis2018reset, gupta2021reset,smith2021legged, ha2020learning, bloesch2022towards}.

\textbf{Distribution Matching in RL.} Critical to our method is matching the state distribution of the demonstrations.
Such a distribution matching perspective is often employed in inverse RL \citep{ng2000algorithms,ziebart2008maximum, ziebart2010modeling,finn2016guided} and imitation learning \citep{ghasemipour2020divergence, argall2009survey} or to encourage efficient exploration \citep{lee2019efficient}. More recently, several works have leveraged implicit distribution matching by posing a classification problem, pioneered in \citet{goodfellow2014generative}, to imitate demonstrations \citep{ho2016generative,baram2017end,kostrikov2018discriminator,rafailov2021visual}, to imitate sequences of observations~\cite{torabi2019adversarial,zhu2020off}, or to learn reward functions for goal-reaching \citep{fu2018variational, singh2019end}. Our work employs a similar discriminator-based approach to encourage the state distribution induced by the policy to match that of the demonstrations. Importantly, our work focuses on creating an initial state distribution that the forward policy can learn efficiently from, as opposed to these prior works that are designed for the episodic RL setting. As the experiments in Section~\ref{sec:gail} and Section~\ref{sec:init_state_ablation} show, na\"ive extensions of these methods to non-episodic settings don't fare well.

\textbf{Accelerating RL using Demonstrations.} There is rich literature on using demonstrations to speed up reinforcement learning, especially for sparse reward problems.
Prior works have considering shaping rewards using demonstrations~\citep{brys2015reinforcement}, pre-training the policy~\citep{rajeswaran2017learning}, using behavior cloning loss as a regularizer for policy gradients~\citep{rajeswaran2017learning} and $Q$-learning~\citep{nair2018overcoming}, and initializing the replay buffer~\citep{nair2018overcoming, vecerik2017leveraging, hester2018deep}. MEDAL leverages demonstrations to accelerate non-episodic reinforcement learning by utilizing demo distribution to create initial conditions for the forward policy. The techniques proposed in these prior works are complimentary to our proposal, and can be leveraged for non-episodic RL in general as well. Indeed, for all methods in our experiments, the replay buffer is initialized with demonstrations.

\section{Preliminaries}
\label{sec:prelims}
\textbf{Autonomous Reinforcement Learning}. We use the ARL framework for non-episodic learning defined in \citet{sharma2021autonomous}, which we briefly summarize here.
Consider a Markov decision process ${\mathcal{M} \equiv (\mathcal{S}, \mathcal{A}, p, r, \rho_0)}$, where $\mathcal{S}$ denotes the state space, $\mathcal{A}$ denotes the action space, ${p : \mathcal{S} \times \mathcal{A} \times \mathcal{S} \mapsto \mathbb{R}_{\geq 0}}$ denotes the transition dynamics, ${r: \mathcal{S} \times \mathcal{A} \mapsto \mathbb{R}}$ denotes the reward function and $\rho_0$ denotes the initial state distribution. The learning algorithm $\mathbb{A}$ is defined as ${\mathbb{A}: \{s_i, a_i, s_{i+1}, r_i\}_{i=0}^t \mapsto \{a_t, \pi_t\}}$, which maps the transitions collected in the environment until time $t$ to an action $a_t$ and its best guess at the optimal policy ${\pi_t: \mathcal{S} \times \mathcal{A} \mapsto \mathbb{R}_{\geq 0}}$. First, the initial state is sampled exactly once ($s_0 \sim \rho_0$) at the beginning of the interaction and the learning algorithm interacts with the environment through the actions $a_t$ till $t \to \infty$. This is the key distinction from an episodic RL setting where the environment resets to a state from the initial state distribution after a few steps. Second, the action taken in the environment does not necessarily come from $\pi_t$, for example, a backward policy $\pi_b$ may generate the action taken in the environment.

ARL defines two metrics: \textit{Continuing Policy Evaluation} measures the reward accumulated by $\mathbb{A}$ over the course of training, defined as ${\mathbb{C}(\mathbb{A}) = \lim_{h \to\infty}\frac{1}{h} \mathbb{E}\left[\sum_{t=0}^h r(s_t, a_t) \right]}$ and
 \textit{Deployed Policy Evaluation} metric measures how quickly an algorithm improves the output policy $\pi_t$ at the task defined by the reward function $r$, defined as:
\begin{equation}
    \mathbb{D}(\mathbb{A}) = \sum_{t=0}^\infty J(\pi^*) - J(\pi_t),
\end{equation}
where ${J(\pi) = \mathbb{E}\left[\sum_{j=0}^\infty \gamma^j r(s_j, a_j) \right], s_0 \sim \rho_0, a_t \sim \pi(\cdot \mid s_t)}$, ${s_{t+1}\sim p(\cdot \mid s_t, a_t)}$ and ${\pi^* \in \argmax_\pi J(\pi)}$. The goal for an algorithm $\mathbb{A}$ is to minimize $\mathbb{D}(\mathbb{A})$, that is to bring $J(\pi_t)$ close to $J(\pi^*)$ in the least number of samples possible. Intuitively, minimizing $\mathbb{D}(\mathbb{A})$ corresponds to maximizing the area under the curve for $J(\pi_t)$ versus $t$.

$\mathbb{C}(\mathbb{A})$ corresponds to the more conventional average-reward reinforcement learning. While algorithms are able to accumulate large rewards during training, they do not necessarily recover the optimal policy in non-episodic settings~\citep{zhu20ingredients, co2020ecological, sharma2021autonomous}. In response, \citet{sharma2021autonomous} introduce $\mathbb{D}(\mathbb{A})$ to explicitly encourage algorithms to learn task-solving behaviors and not just accumulate reward through training.

\textbf{Imitation Learning via Distribution Matching.} Generative Adversarial Networks~\citep{goodfellow2016nips} pioneered implicit distribution matching for distributions where likelihood cannot be computed explicitly. Given a dataset of samples $\{x_i\}_{i=1}^N$, where $x_i \sim p^*(\cdot)$ for some target distribution $p^*$ over the data space $\mathcal{X}$, generative distribution $p_\theta(\cdot)$ can be learned through the following minimax optimization:
\begin{equation}
    \min_{p_\theta} \max_{D} \mathbb{E}_{x \sim p^*} \left[\log D(x)\right] + \mathbb{E}_{x \sim p_\theta} \left[\log(1-D(x)) \right]
\end{equation}
where $D: \mathcal{X} \mapsto [0, 1]$ is discriminator solving a binary classification problem. This can be shown to minimize the Jensen-Shannon divergence, that is $\mathcal{D}_{\textrm{JS}}(p_\theta \mid\mid p^*)$~\citep{goodfellow2014generative,nowozin2016f} by observing that the Bayes-optimal classifier satisfies $D^*(x) = \frac{p^*(x)}{p^*(x) + p^\theta(x)}$ (assuming that prior probability of true data and fake data is balanced). Because we do not require an explicit density under the generative distribution and only require the ability to sample the distribution, this allows construction of imitation learning methods such as GAIL~\cite{ho2016generative} which minimizes $\mathcal{D}_{\textrm{JS}}(\rho^\pi(s, a) \mid\mid \rho^*(s, a))$, where the policy $\pi$ is rolled out in the environment starting from initial state distribution $\rho_0$ to generate samples from the state-action distribution $\rho^\pi(s, a)$ and $\rho^*(s, a)$ is the target state-action distribution of the demonstrations.

\section{Matching Expert Distributions for Autonomous Learning (MEDAL)}
\label{sec:medal}

Several prior works demonstrate the ineffectiveness of standard RL methods in non-episodic settings \citep{co2020ecological,zhu20ingredients,sharma2021autonomous}. Adding noise to actions, for example $\epsilon$-greedy in DQN \citep{mnih2015human} or Gaussian noise in SAC \citep{haarnoja2018soft}), can be sufficient for exploration in episodic setting where every trial starts from a narrow initial state distribution. However, such an approach becomes ineffective in non-episodic settings because the same policy is expected to both solve the task and be sufficiently exploratory. As a result, a common solution in non-episodic autonomous RL settings is to learn another policy in addition to the forward policy $\pi_f$ that solves the task \citep{han2015learning, eysenbach2017leave, zhu20ingredients}:
a backward policy $\pi_b$ that targets a set of states to explore solving the task from. More precisely, the forward policy $\pi_f$ learns to solve the task from a state sampled from $\rho^b$, the marginal state distribution of $\pi^b$.
An appropriate $\rho^b$ can improve the efficiency of learning $\pi_f$ by creating an effective initial state distribution for it. What should the $\pi_b$ optimize? We discuss this question in Section~\ref{sec:kakade} and a practical way to optimize the suggested objective in Section~\ref{sec:disc}. An overview of our proposed algorithm is given in Section~\ref{alg:summary}.

\begin{figure}[!h]
    \centering
    \includegraphics[width=0.8\columnwidth]{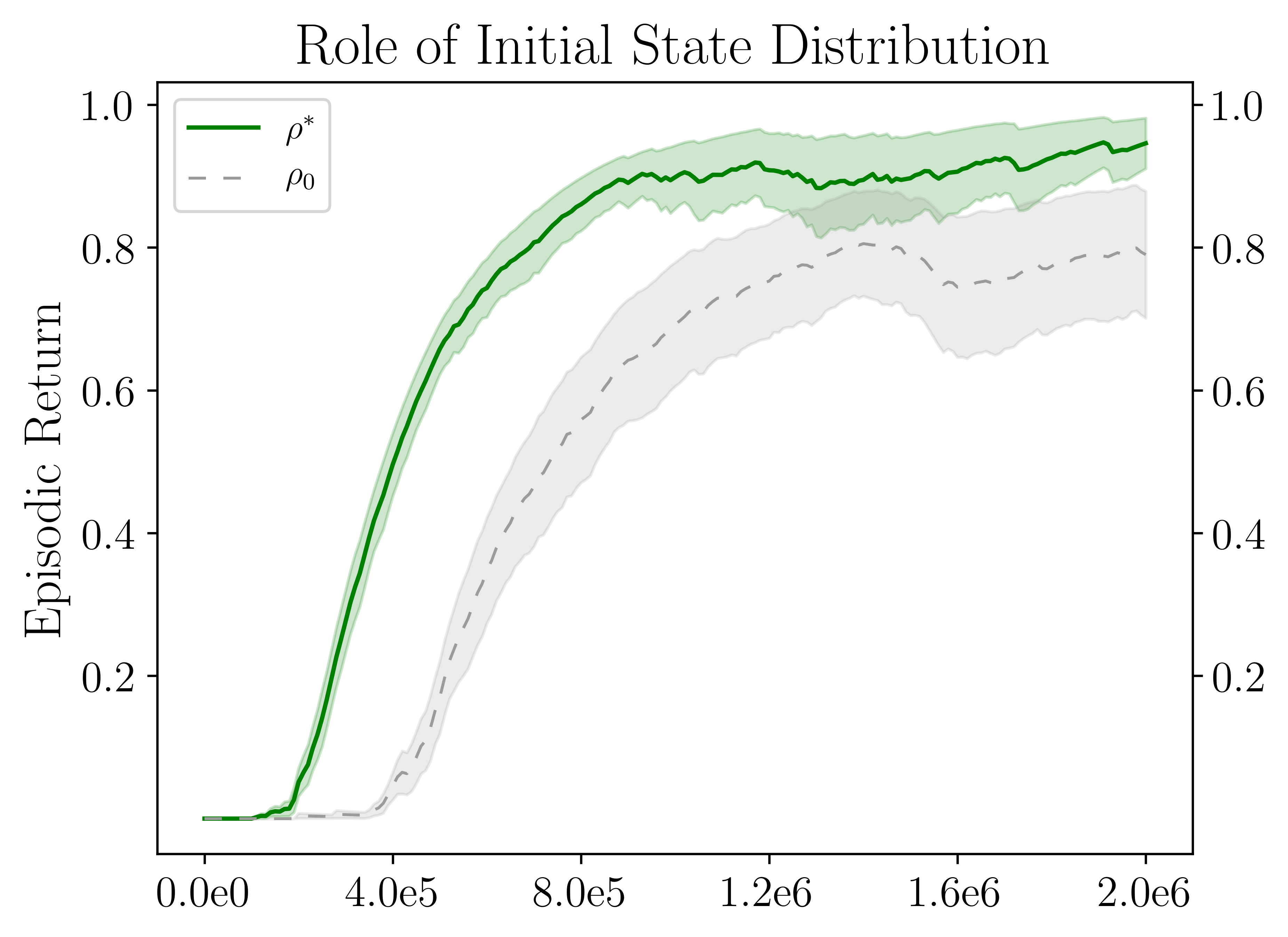}

    \caption{Comparison of sampling initial states $s_0$ from the state distribution of the optimal policy $\rho^*$, with sampling the initial state from the default distribution $\rho_0$ in the episodic setting. The episodic return is computed by initializing the agent at $s_0 \sim \rho_0$ in both the cases. The former improves both the sample efficiency and the performance of the final policy.}
    \label{fig:init_state}
\end{figure}

\subsection{Finding Better Starting States}
\label{sec:kakade}
In episodic settings, $\pi_f$ always starts exploring from $\rho_0$, which is the same distribution from which it will be evaluated. A natural objective for $\pi_b$ then is to minimize the distance between $\rho^b$ and $\rho_0$. And indeed, prior works have proposed this approach \citep{han2015learning, eysenbach2017leave} by learning a backward controller to retrieve the initial state distribution $\rho_0$. While the initial state distribution cannot be changed in the episodic setting, $\pi_b$ does not have any restriction to match $\rho_0$ in the autonomous RL setting. Is there a better initial state distribution to efficiently learn $\pi_f$ from?

Interestingly, \citet{kakade2002approximately} provide a theoretical discussion on how the initial state distribution affects the performance. The main idea is that learning an optimal policy often requires policy improvement at states that are unlikely to be visited. Creating a more uniform starting state distribution can accelerate policy improvement by encouraging policy improvement at those unlikely states. The formal statement can be found in \citep[Corollary~4.5]{kakade2002approximately}. Informally, the result states that the upper bound on the difference between the optimal performance and that of policy $\pi$ is proportional to $\lVert\frac{\rho^*(s)}{\mu}\rVert_{\infty}$, where $\rho^*$ is the state distribution of the optimal policy and $\mu$ is the initial state distribution. This suggests that an initial state distribution $\mu$ that is close to the optimal state distribution $\rho^*$ would enable efficient learning. Intuitively, some initial states in the optimal state distribution would simplify the exploration by being closer to high reward states, which can be bootstrapped upon to learn faster from the harder initial states. To empirically verify the theoretical results, we compare the learning speed of RL algorithm in the episodic setting on \textit{tabletop organization} (environment details in Section~\ref{sec:experiments}) when starting from (a) the standard initial state distribution, that is $s_0 \sim \rho_0$, versus (b) states sampled from the stationary distribution of the optimal policy, that is $s_0 \sim \rho^*(s)$. We find in Figure~\ref{fig:init_state} that the latter not only improves the learning speed, but also improves the performance by nearly 18\%.

\subsection{Resetting to Match the Expert State Distribution}
\label{sec:disc}
The theoretical and empirical results in the previous section suggest that $\pi_f$ should attempt to solve the task from an initial state distribution that is close to $\rho^*(s)$, thus implying that $\pi_b$ should try to match $\rho^*(s)$. 
How do we match $\rho^b$ to $\rho^*$? We will assume access to a small set of samples from $\rho^*(s)$ in the form of demonstrations $\mathcal{D}_f$. Because we are limited to sampling $\rho^b$ and only have a fixed set of samples from $\rho^*$, we consider the following optimization problem:
\begin{equation}
    \label{eq:minimax}
    \min_{\pi_b} \max_C \mathbb{E}_{s \sim \rho^*}\big[\log C(s)\big] + \mathbb{E}_{s \sim \rho^b}\big[\log (1-C(s))\big]
\end{equation}
where $C: \mathcal{S} \mapsto [0, 1]$ is a state-space classifier. This optimization is very much reminiscent of implicit distribution matching techniques used in \citep{goodfellow2014generative, nowozin2016f, ho2016generative, ghasemipour2020divergence} when only the samples are available and densities cannot be explicitly measured.
This can be interpreted as minimizing the Jensen-Shannon divergence $\mathcal{D}_{\textrm{JS}}(\rho^b \mid\mid \rho^*)$.
Following these prior works, $C(s)$ solves a binary classification where $s \sim \rho^*$ has a label $1$ and $s \sim \rho^b$ has a label $0$. Further, $\pi_b$ solves a RL problem to maximize ${\mathbb{E}_{s \sim \rho^b} [r(s, a)]}$, where the reward function $r(s, a) = -\log(1 - C(s))$. Assuming sufficiently expressive non-parametric function classes, $(\rho^*, 0.5)$ is a saddle point for the optimization in Equation~\ref{eq:minimax}.

\textbf{Relationship to Prior Imitation Learning Methods.} GAIL \citep{ho2016generative} proposes to match the state-action distribution $\rho^\pi(s, a)$ to that of the expert $\rho^*(s, a)$, that is minimize $\mathcal{D}_{\textrm{JS}}(\rho^\pi(s, a) \mid\mid \rho^*(s, a))$. Prior works have considered the problem of imitation learning when state-only observations are available \citep{torabi2019adversarial, zhu2020off} by minimizing $\mathcal{D}_{f}(\rho^\pi(s, s') \mid\mid \rho^*(s, s'))$, where $f$-divergence enables generalized treatment of different discrepancy measures such KL-divergence of JS-divergence used in prior work~\citep{nowozin2016f}.
In contrast to these works, our work proposes to minimize $\mathcal{D}_{\textrm{JS}}(\rho^\pi (s) \mid\mid \rho^*(s))$. Furthermore, state distribution matching is only used for the backward policy in our algorithm, whereas the forward policy is maximizing return, as we summarize in the next section. Finally, unlike prior works, the motivation for matching the state distributions is to create an effective initial state distribution for the forward policy $\pi_f$. Our experimental results in Section~\ref{sec:gail} suggest that naively extending GAIL to non-episodic settings is not effective, validating the importance of these key differences.

\begin{algorithm}
    \caption{Matching Expert Distributions for Autonomous Learning (MEDAL)}\label{alg:main}
    \begin{algorithmic}
        \STATE \textbf{require:} forward demos $\mathcal{D}_f$;
        \STATE \textbf{optional:}  backward demos $\mathcal{D}_b$;
        \STATE \textbf{initialize:} $\mathcal{R}_f, \pi_f(a \mid s), Q^{\pi_f}(s, a)$; {\color{olive}// forward policy}
        \STATE \textbf{initialize:} $\mathcal{R}_b, \pi_b(a \mid s), Q^{\pi_b}(s, a)$; {\color{olive}// backward policy}
        \STATE \textbf{initialize:} $C(s)$; {\color{olive}// state-space discriminator}
        \STATE $\mathcal{R}_f \gets \mathcal{R}_f \cup \mathcal{D}_f, \mathcal{R}_b \gets \mathcal{R}_b \cup \mathcal{D}_b$;
        \STATE $s \sim \rho_0$; {\color{olive} // sample initial state}
        \WHILE {not done}
            \STATE {\color{olive}// run forward policy for a fixed number of steps or until goal is reached, otherwise run backward policy}
            \IF{forward}
                \STATE $a \sim \pi_f( \cdot \mid s)$;
                \STATE $s' \sim p(\cdot \mid s, a), r \leftarrow r(s, a)$;
                \STATE $\mathcal{R}_f \gets \mathcal{R}_f \cup \{(s, a, s', r)\}$;
                \STATE update $\pi_f, Q^{\pi_f}$;
            \ELSE
                \STATE $a \sim \pi_b( \cdot \mid s)$;
                \STATE $s' \sim p(\cdot \mid s, a), r \leftarrow -\log (1 - C(s'))$;
                \STATE $\mathcal{R}_b \gets \mathcal{R}_b \cup \{(s, a, s', r)\}$;
                \STATE update $\pi_b, Q^{\pi_b}$;
            \ENDIF
            \STATE {\color{olive} // train disriminator every $K$ steps}
            \IF {train-discriminator}
                \STATE {\color{olive} // sample a batch of positives $S_p$ from the forward demos $\mathcal{D}_f$, and a batch of negatives $S_n$ from backward replay buffer $\mathcal{R}_b$}
                \STATE $S_p \sim \mathcal{D}_f, S_n \sim \mathcal{R}_b$;
                \STATE update $C$ on $S_p \cup S_n$;
            \ENDIF
            \STATE $s \gets s'$;
        \ENDWHILE
    \end{algorithmic}
\end{algorithm}

\subsection{MEDAL Overview}
\label{alg:summary}
With these components in place, we now summarize our proposed algorithm, \emph{Matching Expert Distributions for Autonomous Learning} (MEDAL). We simultaneously learn the following components: a \emph{forward policy} that learns to solve the task and will also be used for evaluation, a \textit{backward policy} that learns creates the initial state distribution for the forward policy by matching the state distribution in the demonstrations, and finally a \textit{state-space discriminator} that learns to distinguish between the states visited by the backward policy and the states visited in the demonstrations. MEDAL assumes access to a set of forward demonstrations $\mathcal{D}_f$, completing the task from the initial state distribution, and optionally, a set of backward demonstrations $\mathcal{D}_b$ undoing the task back to the initial state distribution. The forward policy $\pi_f$ is trained to maximize $\mathbb{E}[{\sum_{t=0}^\infty \gamma^t r(s_t, a_t)}]$ and the replay buffer for the forward policy is initialized using $\mathcal{D}_f$. The backward policy $\pi_b$ trains to minimize ${\mathcal{D}_{\textrm{JS}}(\rho^b(s) \mid\mid \rho^*(s))}$ which translates into maximizing ${-\mathbb{E}[\sum_{t=0}^\infty \gamma^t \log(1-C(s_{t+1}))]}$ and the replay buffer for the backward policy is initialized using the backward demonstrations $\mathcal{D}_b$, if available. Finally, the state-space discriminator $C(s)$ trains to classify states sampled from the \textit{forward} demonstrations $\mathcal{D}_f$ with label $1$ and states visited by $\pi_b$ as label $0$. Note, we are trying to match the state marginal of policy $\pi_b$ (i.e. $\rho^b(s)$) to the optimal state distribution $\rho^*(s)$ (approximated via \textit{forward demonstrations} $\mathcal{D}_f$, not \textit{backward demonstrations}), thereby motivating the classification problem for $C(s)$.

When interacting with the environment during training, we alternate between collecting samples using $\pi_f$ for a fixed number of steps and collecting samples using $\pi_b$ for a fixed number of steps. The policies can be updated using any RL algorithm. The state-space discriminator $C(s)$ is updated every $K$ steps collected in the environment, with the states visited by $\pi_b$ being labeled as $0$ and states in $\mathcal{D}_f$ labeled as $1$. The minibatch for updating the parameters of $C(s)$ is balanced to ensure equal samples from $\rho^*(s)$ and $\rho^b(s)$. The pseudocode for MEDAL is provided in Algorithm~\ref{alg:main}, and further implementation details can be found in Appendix~\ref{sec:append_medal}.

\begin{figure*}[t]
    \centering
    \includegraphics[height=0.19\textwidth]{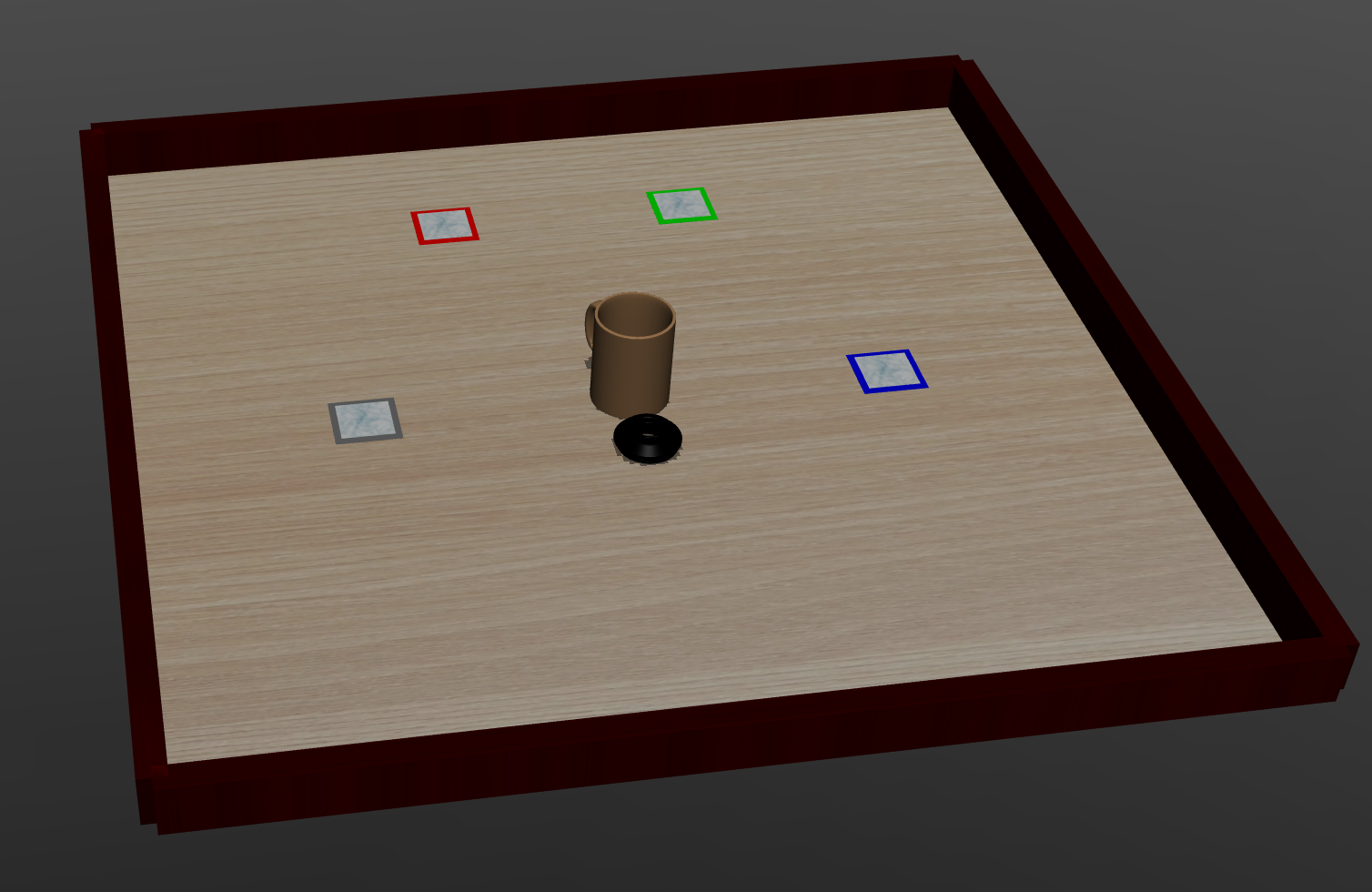}
    \includegraphics[height=0.19\textwidth]{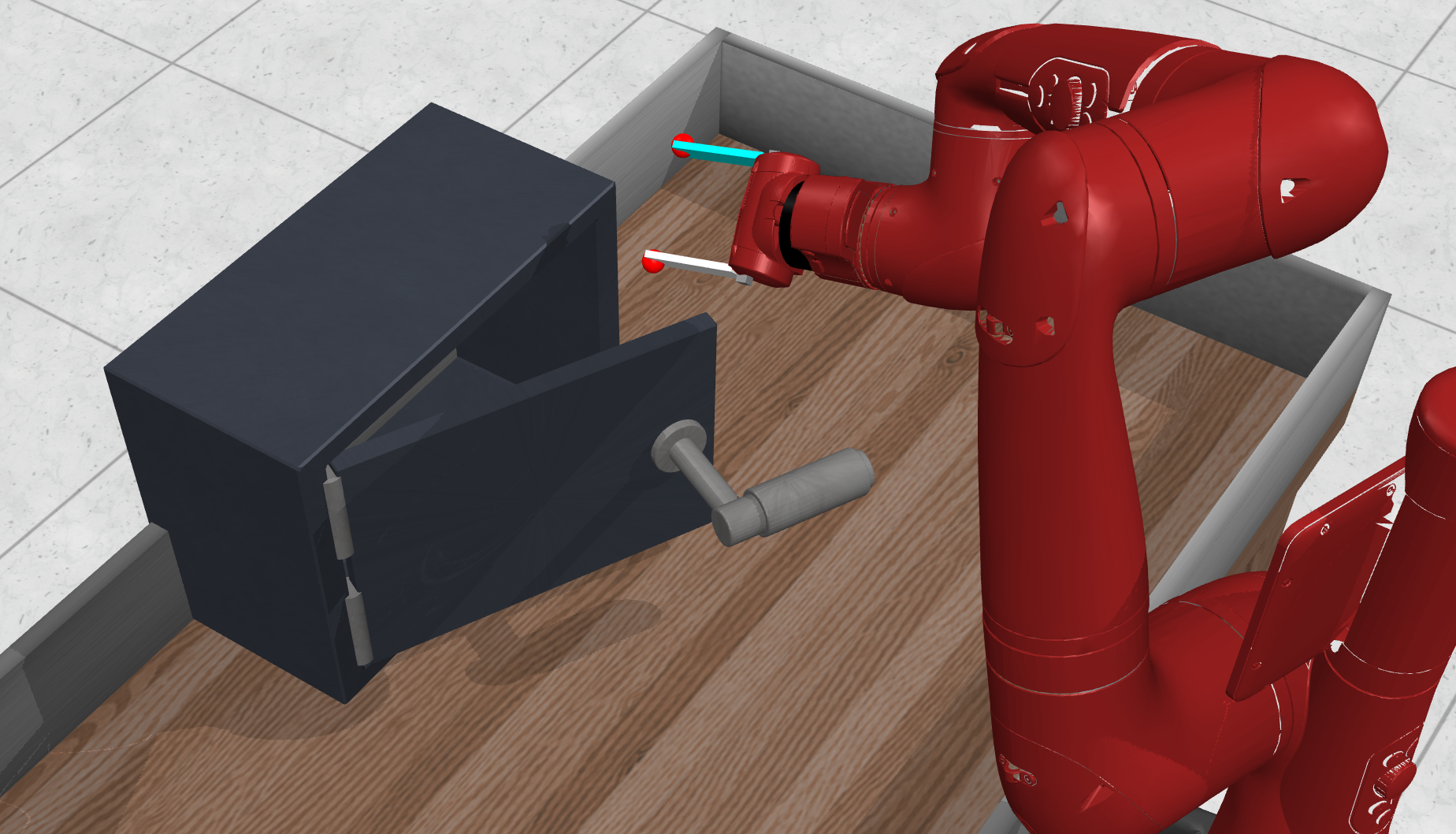}
    \includegraphics[height=0.19\textwidth]{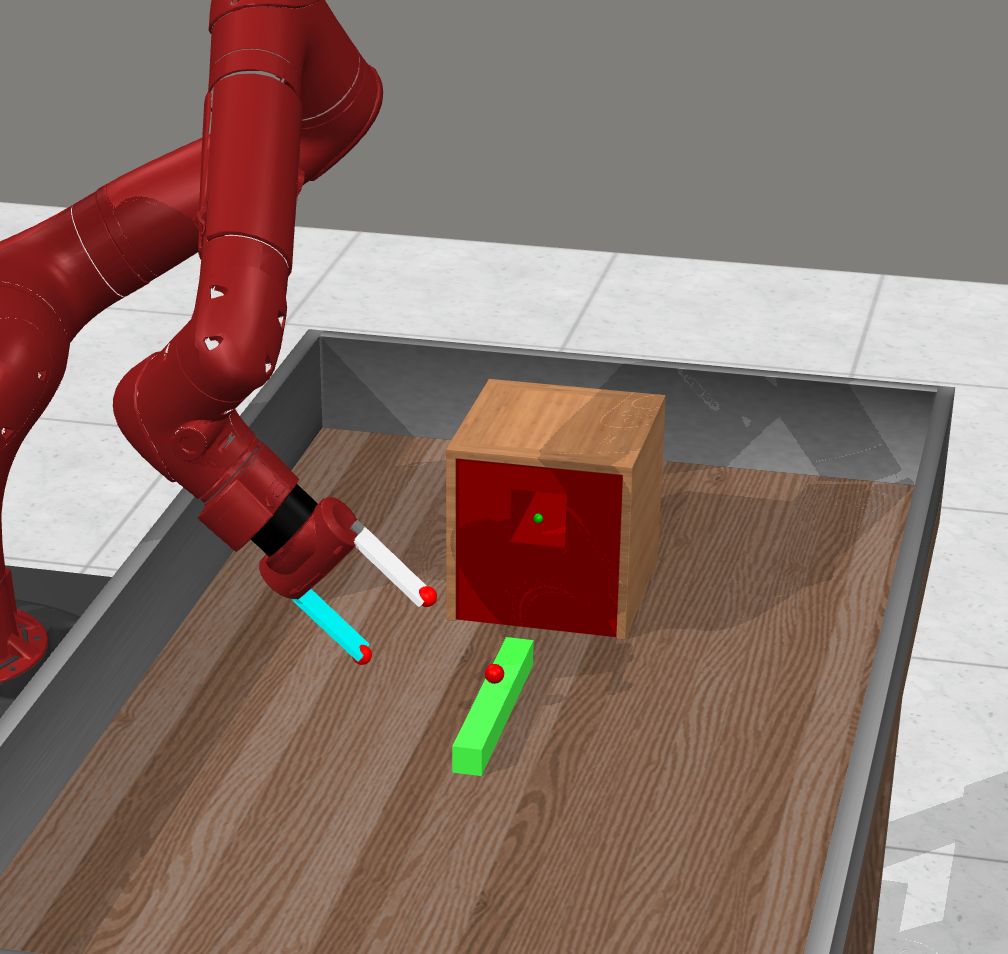}
    \caption{Continuous-control environments from the EARL benchmark: (\emph{left}) Table-top organization (\textbf{TO}) where a gripper is tasked with moving a mug to one of the four goal locations, (\emph{center}) sawyer door closing (\textbf{SD}) where the sawyer robot is tasked with closing the door, (\emph{right}) sawyer peg insertion (\textbf{SP}) where the robot is tasked with picking up the peg and inserting into the goal location.}
    \label{fig:env_images}
\end{figure*}

\section{Experiments}
\label{sec:experiments}

In this section, we empirically analyze the performance of MEDAL to answer to following questions: (1) How does MEDAL compare to other non-episodic, autonomous RL methods? (2) Given the demonstrations, can existing imitation learning methods suffice? (3) How important is it for the backward controller to match the entire state distribution, instead of just the initial state distribution? 

\textbf{Environments.} To analyze these questions, we consider three sparse-reward continuous-control environments from the EARL benchmark \citep{sharma2021autonomous}.
The \emph{table-top organization} is a simplified manipulation environment where a gripper is tasked to move the mug to one of four coasters.
The \emph{sawyer door closing} environment requires a sawyer robot arm to learn how to close a door starting from various initial positions. The challenge in the ARL setting arises from the fact that the agent has to open the door to practice closing it again. Finally, the \emph{sawyer peg insertion} environment requires the sawyer robot arm to pick up a peg and insert it into a designated goal location. This is a particularly challenging environment as the autonomously operating robot can push the peg into places where it can be hard to retrieve it back, a problem that is not encountered in the episodic setting as the environment is reset to the initial state distribution every few hundred steps.

\textbf{Evaluation}. We follow the evaluation protocol laid down in the EARL benchmark. All algorithms are reset to a state $s_0 \sim \rho_0$ and interact with their environments almost fully autonomously thereon, only being reset to an initial state intermittently after several hundreds of thousands of steps of interaction. Since our objective is to acquire task policies in a sample efficient way, we will focus on \textit{deployed policy evaluation}. Specifically, we approximate $J(\pi_t) = \mathbb{E}[\sum_{t=0}^\infty \gamma^t r(s_t, a_t)]$ by averaging the return of the policy over $10$ episodes starting from $s_0 \sim \rho_0$, every $10,000$ steps collected in the training environment. Note, the trajectories collected for evaluation are \textit{not} provided to the learning algorithm $\mathbb{A}$. For all considered environments, the reward functions are sparse in nature and correspondingly, EARL provides a small set of demonstrations to the algorithms, that correspond to doing and undoing the task (a total of $10$-$30$ demonstrations depending on the environment). Environment specific details such as reward functions and intermittent resets can be found in Appendix~\ref{sec:append_environment}.

\subsection{Benchmarking MEDAL on EARL}
\label{sec:benchmark}

\begin{table}[t]
    \label{table:transfer results}
    \centering
    \begin{tabular}{@{}c|ccc@{}}
     \toprule
     \textbf{Method} & \textbf{Tabletop} & \textbf{Sawyer} & \textbf{Sawyer}\\
     & \textbf{Organization} & \textbf{Door} &\textbf{Peg}\\
        \midrule
    \emph{na\"{i}ve RL} & 0.32 (0.17) & 0.00 (0.00) & 0.00 (0.00)\\
    \emph{FBRL}         & 0.94 (0.04) & \textbf{1.00 (0.00)} & 0.00 (0.00)\\
    \emph{R3L}          & 0.96 (0.04) & 0.54 (0.18) & 0.00 (0.00)\\
    \emph{VaPRL}        & \textbf{0.98 (0.02)} & 0.94 (0.05) & 0.00 (0.00)\\
    \emph{MEDAL}        & \textbf{0.98 (0.02)} & \textbf{1.00 (0.00)} & \textbf{0.40 (0.16)}\\
    \hline
    \emph{oracle RL}    & 0.80 (0.11) & 1.00 (0.00) & 1.00 (0.00)\\
    \bottomrule
    \end{tabular}
  \caption{Average return of the final learned policy. Performance is averaged over 5 random seeds. The mean and and the standard error are reported, with the best performing entry in bold. For all domains, $1.0$ indicates the maximum performance and $0.0$ indicates minimum performance.}
\end{table}

First, we benchmark our proposed method MEDAL on the aforementioned EARL environments against state-of-the-art non-episodic ARL methods.

\textbf{Comparisons.} We briefly review the methods benchmarked on EARL, which MEDAL will be compared against: (1) forward-backward RL (\textbf{FBRL}) \citep{han2015learning, eysenbach2017leave}, where the backward policy recovers the initial state distribution; (2) value-accelerated persistent RL (\textbf{VaPRL}) \citep{sharma2021autonomouscurr}, where the backward policy creates a curriculum based on the forward policy's performance; (3) \textbf{R3L} \citep{zhu20ingredients} has a backward policy that optimizes a state-novelty reward \citep{burda2018exploration} to encourage the forward policy to solve the tasks from new states in every trial; (4) \textbf{na\"ive RL} represents the episodic RL approach where only a forward policy optimizes the task-reward throughout training; and finally (5) \textbf{oracle RL} is the same episodic RL baseline but operating in the episodic setting. For a fair comparison, the forward policy for all baselines use SAC~\citep{haarnoja2018soft}, and the replay buffer is always initialized with the forward demonstrations. Further, the replay buffers for backward policies in \textbf{FBRL, VaPRL} is also initialized with the backward demos. The replay buffer of the backward policy in R3L is not initialized with backward demos as it will reduce the novelty of the states in the backward demos for the RND reward without the backward policy ever visiting those states.

\begin{figure*}[t]
    \centering
    \includegraphics[width=0.32\textwidth]{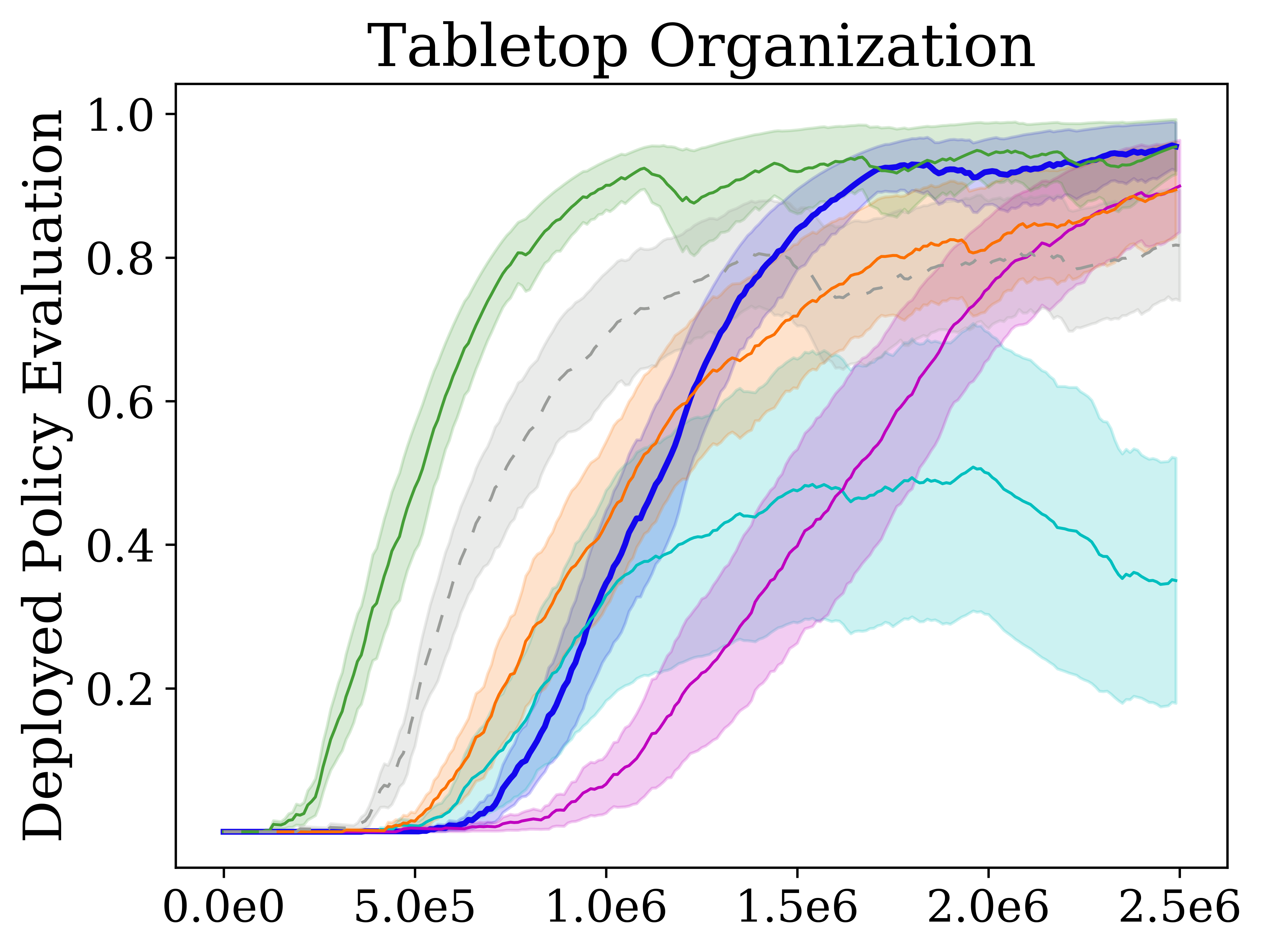}
    \includegraphics[width=0.3\textwidth]{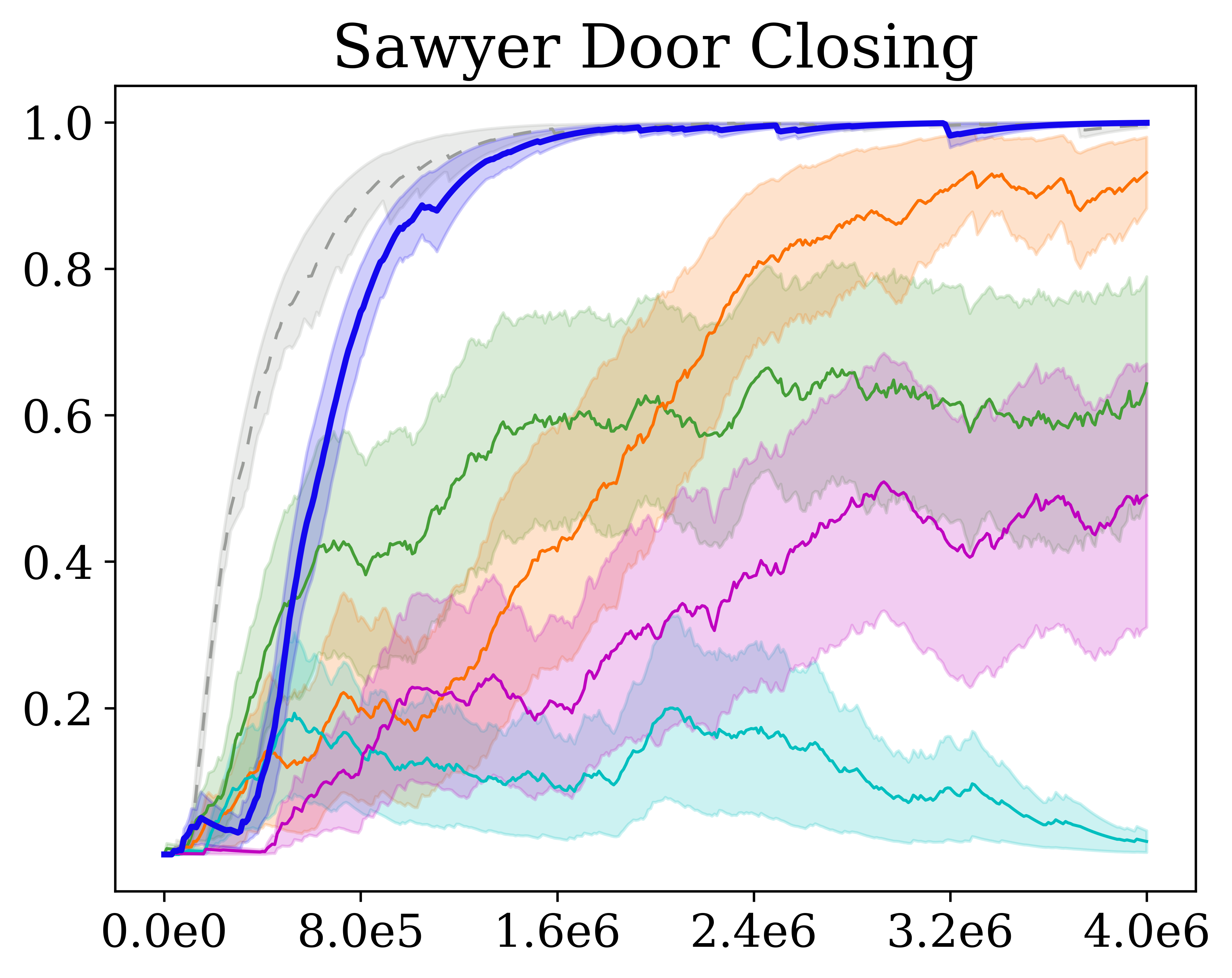}
    \includegraphics[width=0.3\textwidth]{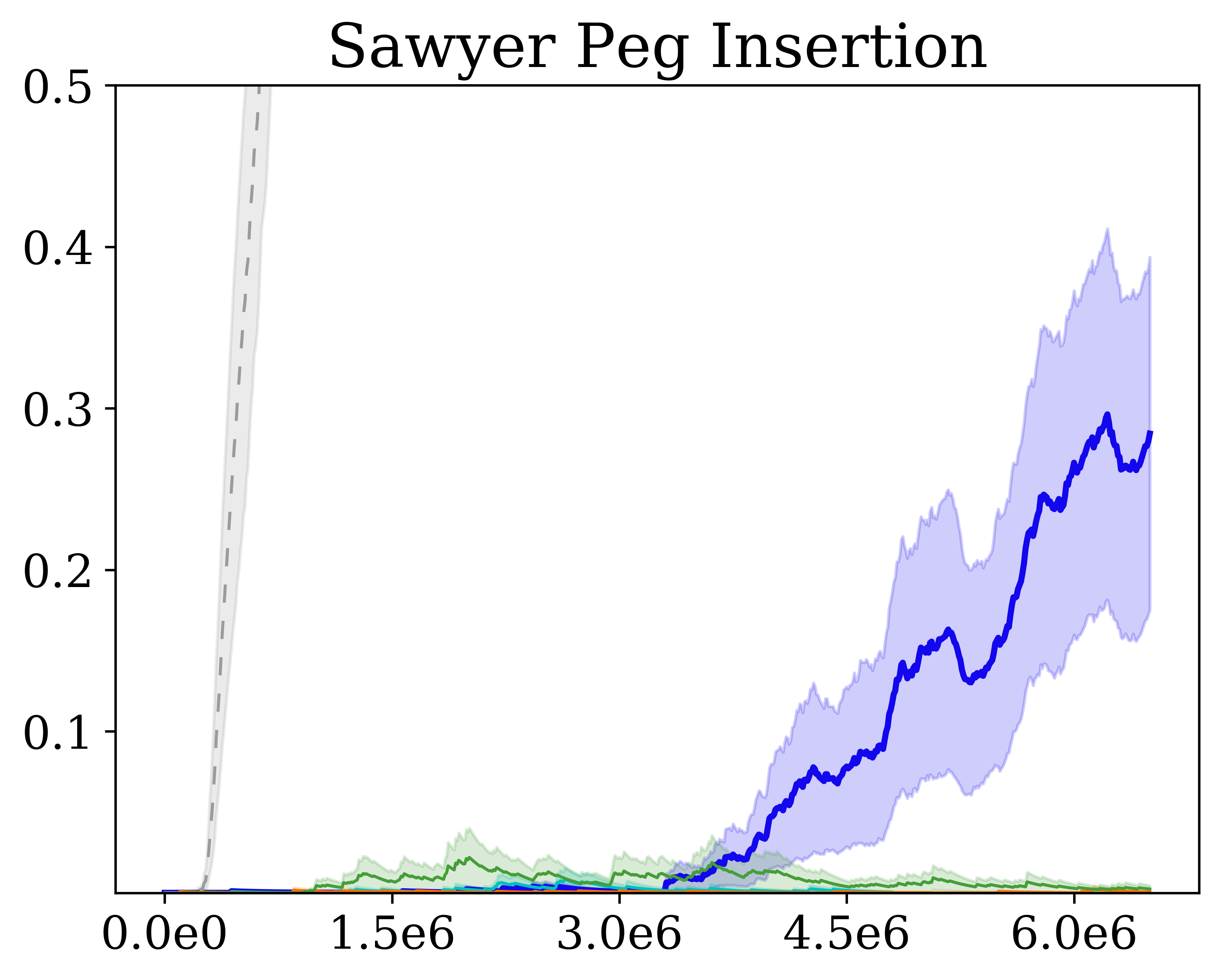}
    \includegraphics[width=0.6\textwidth]{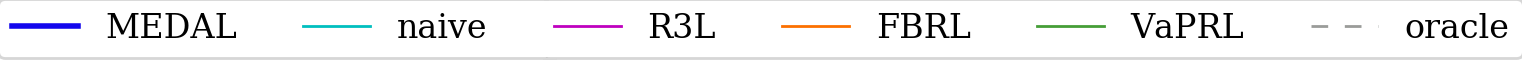}
    
    \caption{Performance of each method on (\textit{left}) the table-top organization environment, (\textit{center}) the sawyer door closing environment, and (\textit{right}) the sawyer peg environment. Plots show learning curves with mean and standard error over 5 random seeds.}
    \label{fig:benchmark_results}
\end{figure*}

It's important to note that some of these comparisons make additional assumptions compared to MEDAL:
\setlist{nolistsep}
\begin{itemize}[noitemsep]
    \item \textbf{oracle RL} operates in the episodic setting, that is the environment is reset to a state from the initial state distribution every few hundred steps. The baseline is included as a reference to compare performance of baselines in ARL versus the conventional episodic setting. It also enables us to compare the performance of conventional RL algorithms when moving from the episodic setting to the ARL setting, by comparing the performance of \textbf{oracle RL} and \textbf{na\"ive RL}.
    \item \textbf{VaPRL} relies on relabeling goals which requires the ability to query the reward function for any arbitrary state and goal (as the VaPRL curriculum can task the agent to reach arbitrary goals from the demonstrations). Additionally, VaPRL has a task specific hyperparameter that controls how quickly the curriculum moves towards the initial state distribution.
\end{itemize}
In a real-world settings, where the reward function often needs to be learned as well (for example from images), these assumptions can be detrimental to their practical application. While FBRL also requires an additional reward function to reach the initial state distribution, the requirement is not as steep. Additionally, we consider a version of FBRL that learns this reward function in Section~\ref{sec:init_state_ablation}. However, the ability to query the reward function for arbitrary states and goals, as is required by VaPRL, can be infeasible in practice. The impact of these additional assumptions cannot be overstated, as the primary motivation for the autonomous RL framework is to be representative of real-world RL training.

\textbf{Results.}
Table~\ref{table:transfer results} shows the performance of the final forward policy, and Figure~\ref{fig:benchmark_results} shows the \textit{deployed} performance of the forward policy versus the training time for different methods.
MEDAL consistently outputs the best performing final policy, as can be seen in Table~\ref{table:transfer results}. Particularly notable is the performance on \textit{sawyer peg insertion}, where the final policy learned by MEDAL gets 40\% success rate on average, while all other methods fail completely. With the exception of VaPRL on \textit{tabletop organization}, MEDAL also learns more efficiently compared to any of the prior methods. Notably, MEDAL substantially reduces the sample efficiency gap between ARL methods and episodic methods on \textit{sawyer door closing}.

We posit two reasons for the success of MEDAL: (a) Learning a backward policy that retrieves the agent close to the task distribution enables efficient exploration, producing the speedup in performance. (b) Bringing the agent closer to the state distribution implicit in the demonstrations may be easier to maximize compared to other objectives, for example, retrieving the agent to the initial state distribution.

\begin{figure}[!h]
    \centering
    \includegraphics[width=0.8\columnwidth]{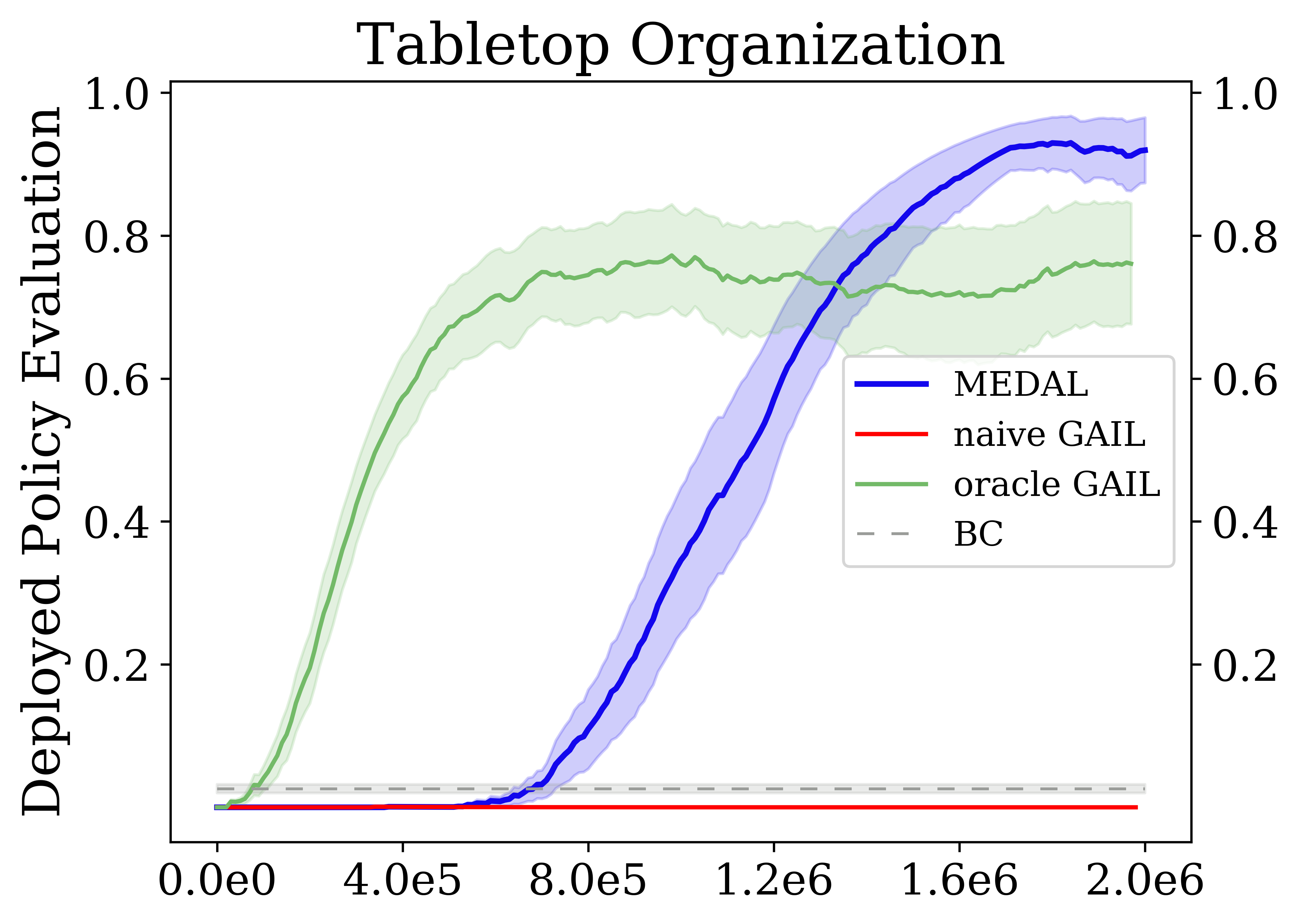}
    \includegraphics[width=0.8\columnwidth]{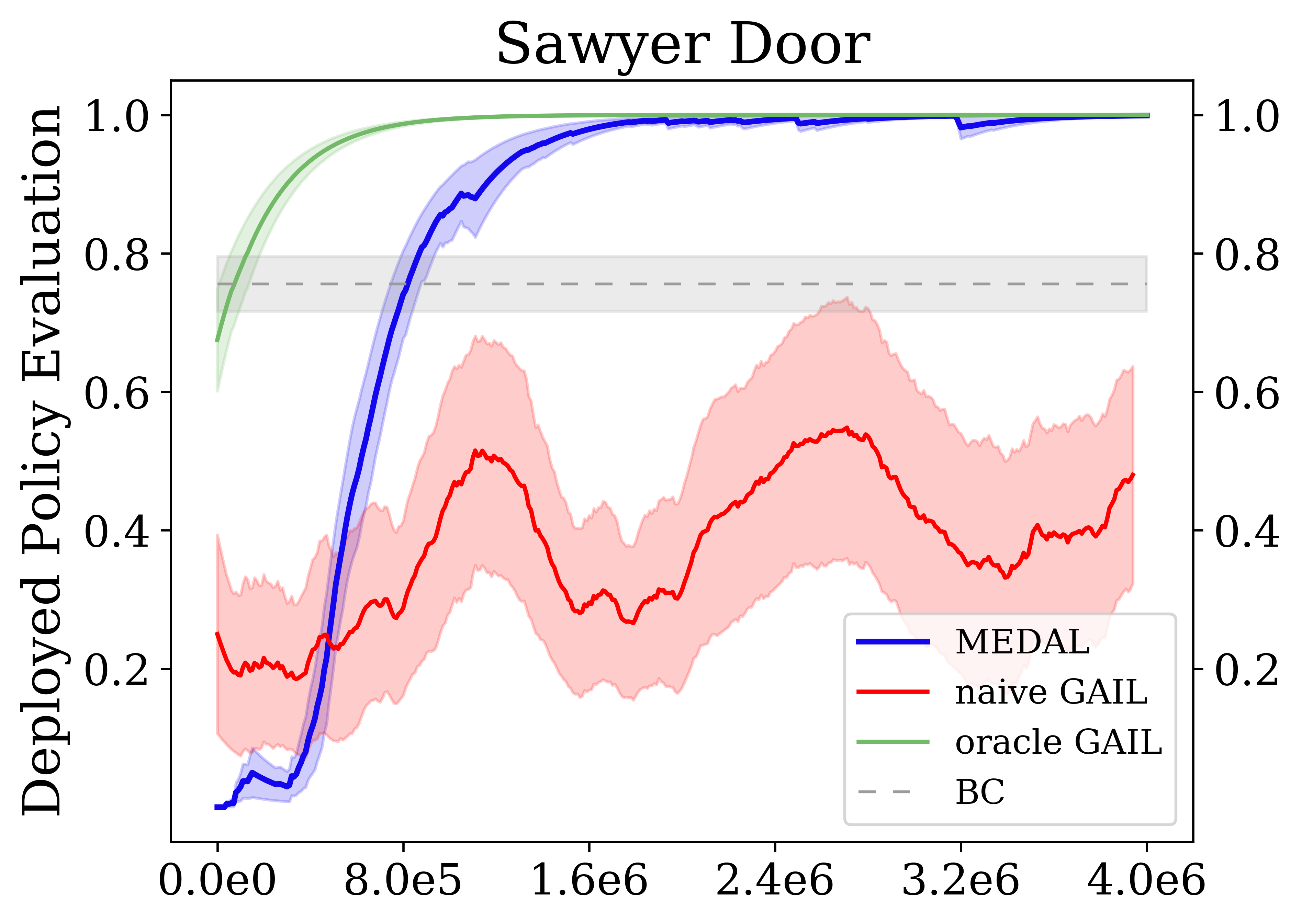}
    \caption{\textbf{MEDAL} in comparison to imitation learning methods on \textit{tabletop organization} and \textit{sawyer door closing}. Behavior cloning (\textbf{BC}) does not fare well, suggesting the importance of online data collection. The success of online imitation learning methods such as GAIL in episodic settings does not translate to the non-episodic ARL setting, as indicated by the substantial drop in performance of \textbf{na\"ive GAIL} compared to \textbf{oracle GAIL}.}
    \label{fig:gail}
\end{figure}

\subsection{Imitation Learning}
\label{sec:gail}
Given that MEDAL assumes access to a set of demonstrations, a natural alternative to consider is imitation learning. In this section, we focus our experiments on the tabletop organization environment. We first test how a behavior cloning fares (\textbf{BC}). Results in Figure~\ref{fig:gail} suggest that behavior cloning does not do well on \textit{tabletop organization}, completely failing to solve the task and leaves substantial room for improvement on \textit{sawyer door}. This is to be expected as EARL provides only a small number of demonstrations, and errors compounding over time from imperfect policies generally leads to poor performance. How do imitation learning methods with online data collection fare? We consider an off-policy version of GAIL, Discriminator Actor-Critic (DAC) \citep{kostrikov2018discriminator}, which matches $\rho^\pi(s, a)$ to $\rho^*(s, a)$ with an implicit distribution matching approach similar to ours. Assuming that $\rho^\pi(s, a)$ can match $\rho^*(s, a)$, the method should in principle recover the optimal policy -- there is nothing specific about GAIL that restricts it to the episodic setting. However, as the results in Figure~\ref{fig:gail} suggest, there is a substantial drop in performance when running GAIL in episodic setting (\textbf{oracle GAIL}) versus the non-episodic ARL setting (\textbf{na\"ive GAIL}). While such a distribution matching could succeed, na\"ively extending the methods to the ARL setting is not as successful, suggesting that it may require an additional policy (similar to the backward policy) to be more effective.

\subsection{The Choice of State Distribution}
\label{sec:init_state_ablation}
\begin{figure}[!h]
    \centering
    \includegraphics[width=0.8\columnwidth]{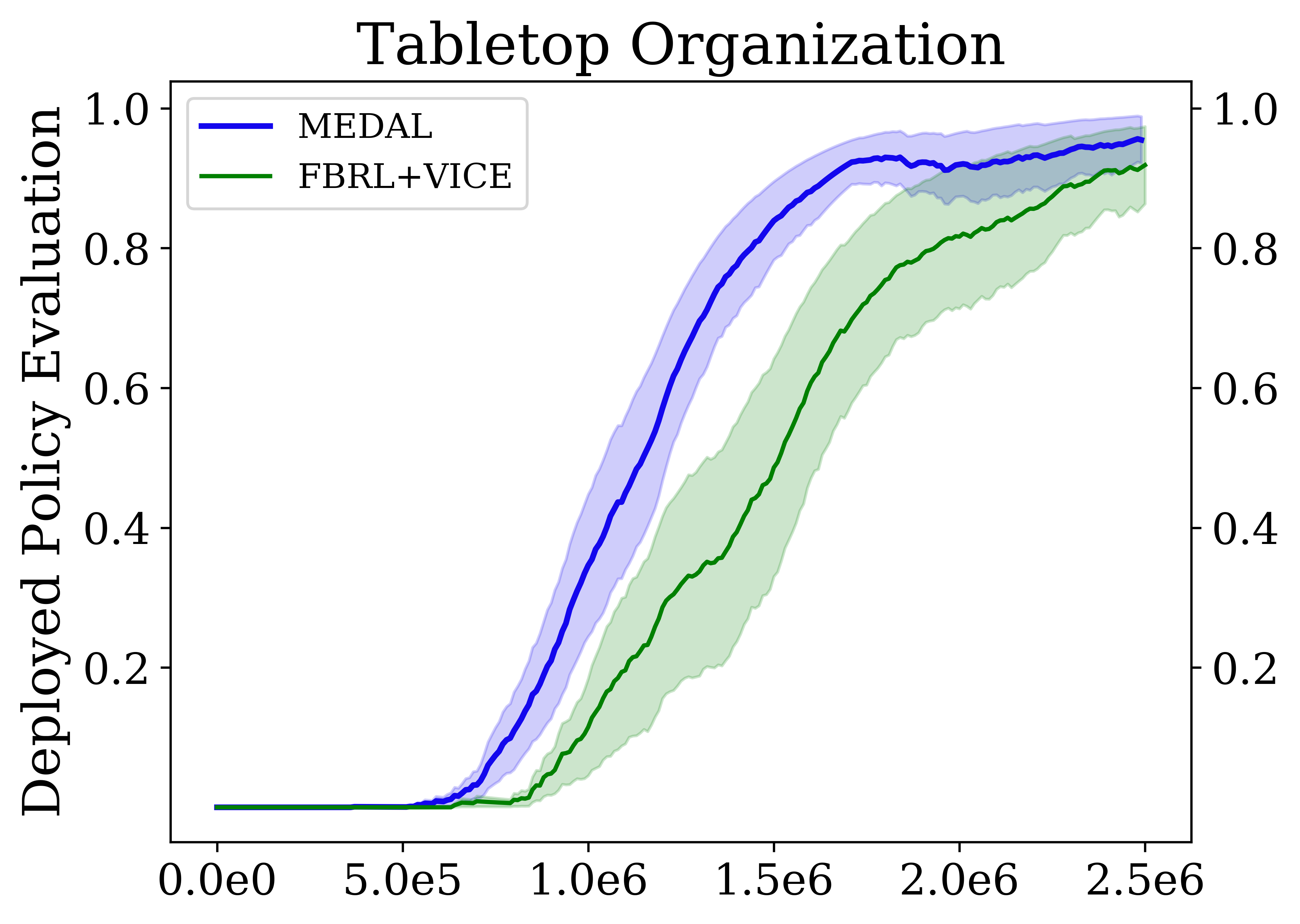}
    \includegraphics[width=0.8\columnwidth]{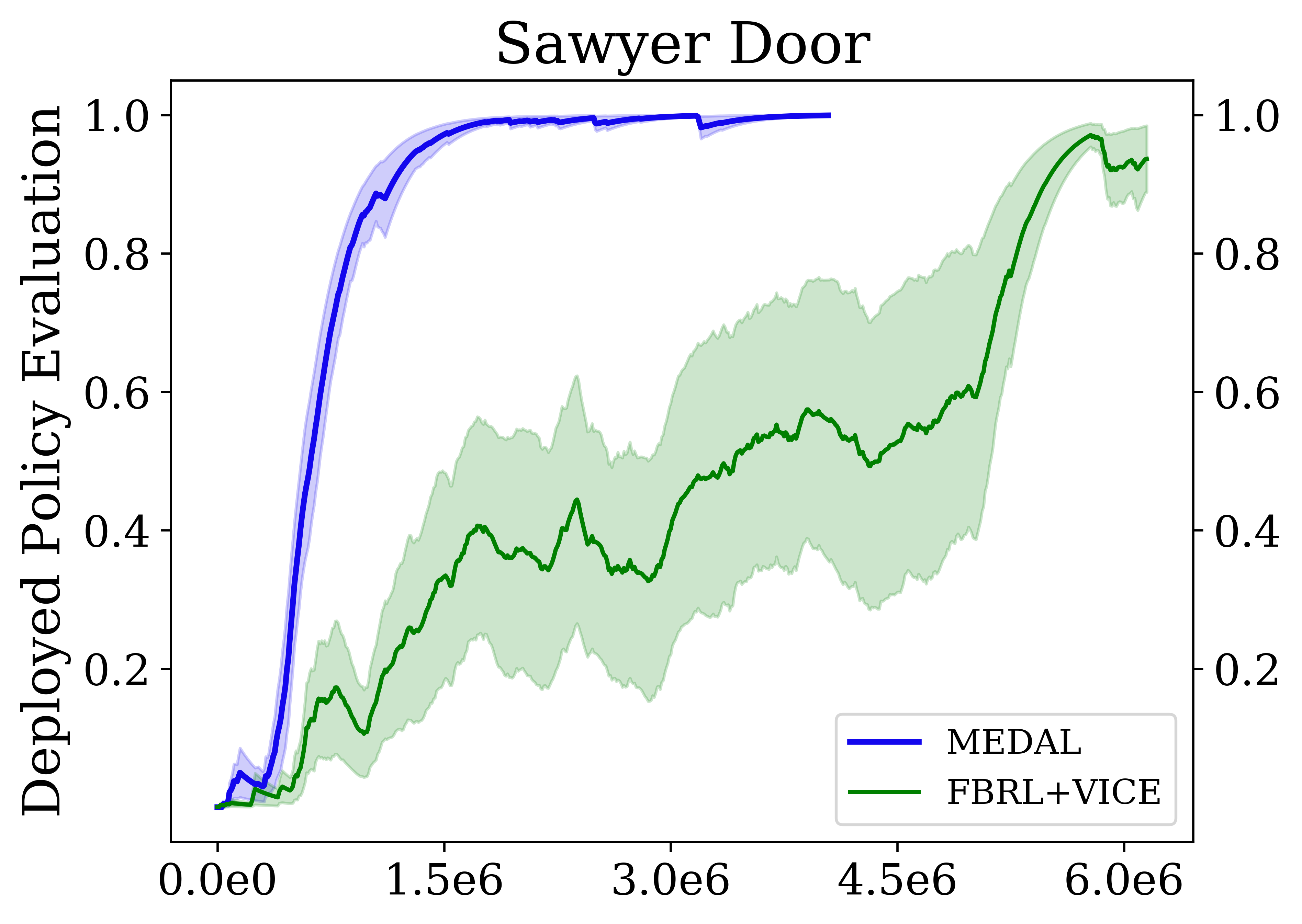}
    \caption{Isolating the effect of matching demonstration data. The speed up of \textbf{MEDAL} compared to \textbf{FBRL + VICE}, which matches the initial state distribution, suggests that the performance gains of MEDAL can be attributed to the better initial state distribution created by the backward controller.}
    \label{fig:vice}
\end{figure}

The key element in MEDAL is matching the state distribution of the backward policy to the states in the demonstrations. To isolate the role of our proposed scheme of minimizing $\mathcal{D}_{\textrm{JS}}(\rho^b \mid\mid \rho^*)$, we compare it to an alternate method that minimizes $\mathcal{D}_{\textrm{JS}} (\rho^b \mid\mid \rho_0)$, i.e., matching the initial state distribution $\rho_0$ instead of $\rho^*$. This makes exactly one change to MEDAL: instead of sampling positives for the discriminator $C(s)$ from forward demonstrations $\mathcal{D}_f$, the positives are sampled from $\rho_0$. Interestingly, this also provides a practically realizable implementation of \textbf{FBRL}, as it removes the requirement of the additional reward function required for the learning a backward policy to reach the initial state distribution.
We call this method \textbf{FBRL + VICE} as VICE~\citep{singh2019end} enables learning a goal reaching reward function using a few samples of the goal distribution, in this case the goal distribution for $\pi_b$ being $\rho_0$. As can be seen in Figure~\ref{fig:vice}, the FBRL + VICE learns slower than MEDAL, highlighting the importance of matching the entire state distribution as done in MEDAL.

\section{Conclusion}
We propose MEDAL, an autonomous RL algorithm that learns a backward policy to match the expert state distribution using an implicit distribution matching approach. Our empirical analysis indicates that this approach creates an effective initial state distribution for the forward policy, improving both the performance and the efficiency. The simplicity of MEDAL also makes it more amenable for the real-world, not requiring access to additional reward functions.

MEDAL assumes access to a (small) set of demonstrations, which may not be feasible in several real-world scenarios. Identifying good initial state distributions without relying on a set of demonstrations would increase the applicability of MEDAL. Similarly, in applications where safe exploration is a requirement, MEDAL can be adapted to constrain the forward policy such that it stays close to the task-distribution defined by the demonstrations. While MEDAL pushes further the improvements in ARL, as exemplified by the reduction of sample efficiency gap on \textit{sawyer door closing} results, there is still a substantial gap in performance between ARL methods and oracle RL on \textit{sawyer peg}, motivating the search for better methods.

\bibliography{example_paper}
\bibliographystyle{icml2022}

\newpage
\appendix
\onecolumn

\section{MEDAL Implementation}
\label{sec:append_medal}
MEDAL is implemented with TF-Agents, built on SAC as the base RL algorithm. Hyperparameters follow the default values: \texttt{initial collect steps}: 10,000, \texttt{batch size} sampled from replay buffer for updating policy and critic: 256, steps \texttt{collected per iteration}: 1, \texttt{trained per iteration}: 1, \texttt{discount factor}: 0.99, \texttt{learning rate}: $3e-4$ (for critics, actors, and discriminator). The actor and critic network were parameterized as neural networks with two hidden layers each of size $256$. For the discriminator, it was parameterized  as a neural network with one hidden layer of size $128$. This discriminator is updated once every 10 collection steps for all environments. Due to a small positive dataset, mixup \citep{zhang2017mixup} is used as a regularization technique on the discriminator for all environments. Additionally, the batch size for the discriminator is set to $800$ for all environments as this significantly larger value was found to stabilize training. 

\begin{figure*}[t]
    \centering
    \includegraphics[width=0.32\textwidth]{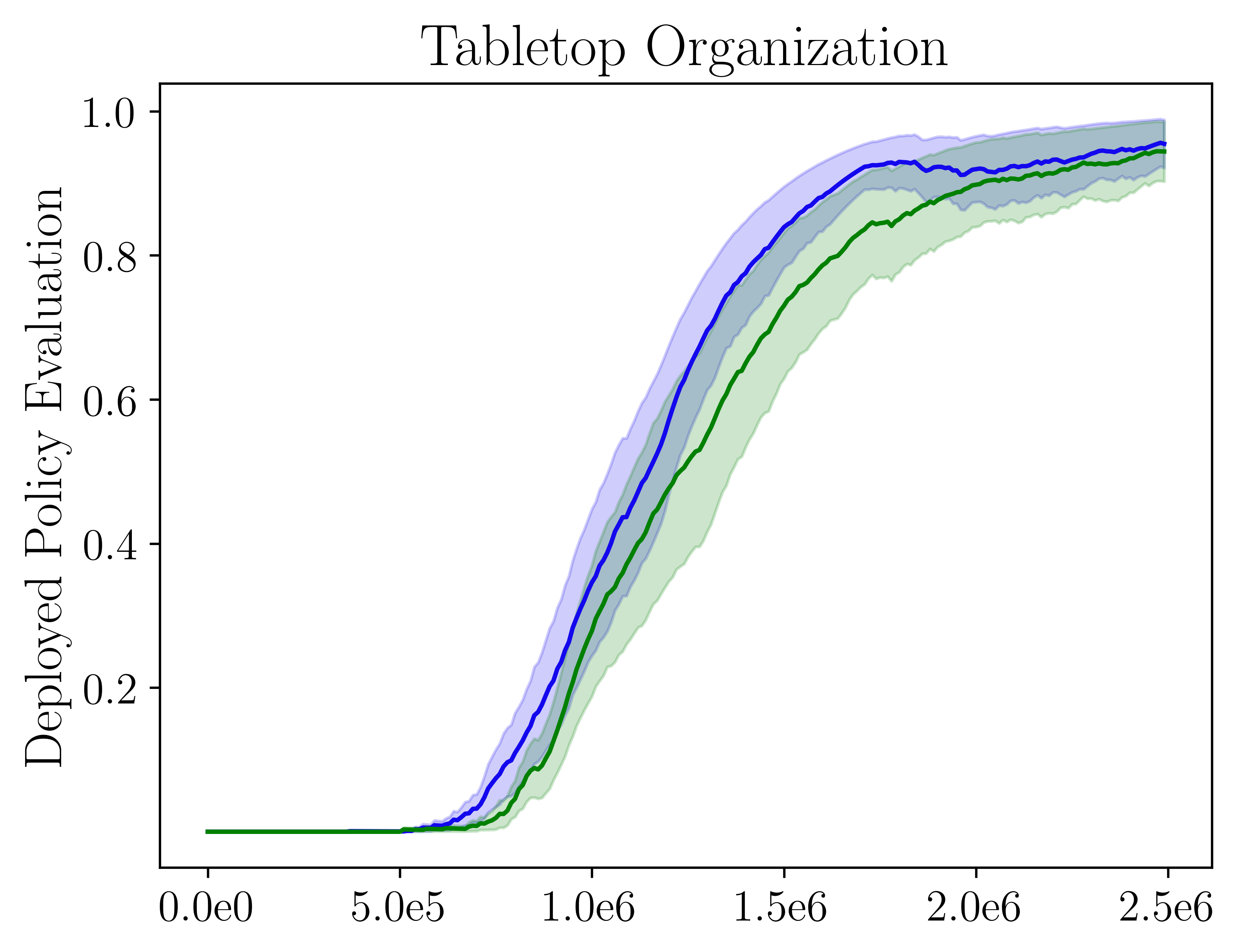}
    \includegraphics[width=0.3\textwidth]{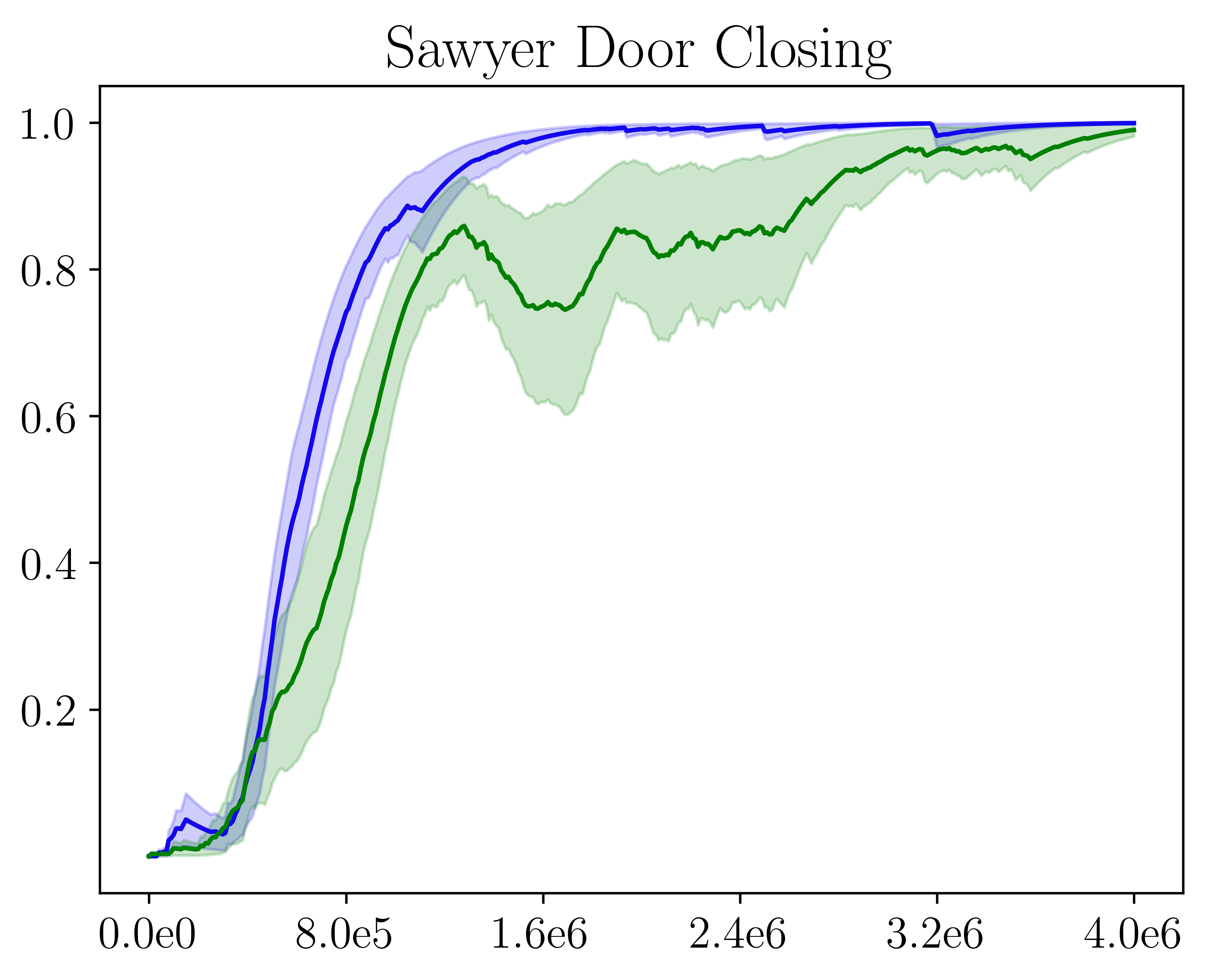}
    \includegraphics[width=0.3\textwidth]{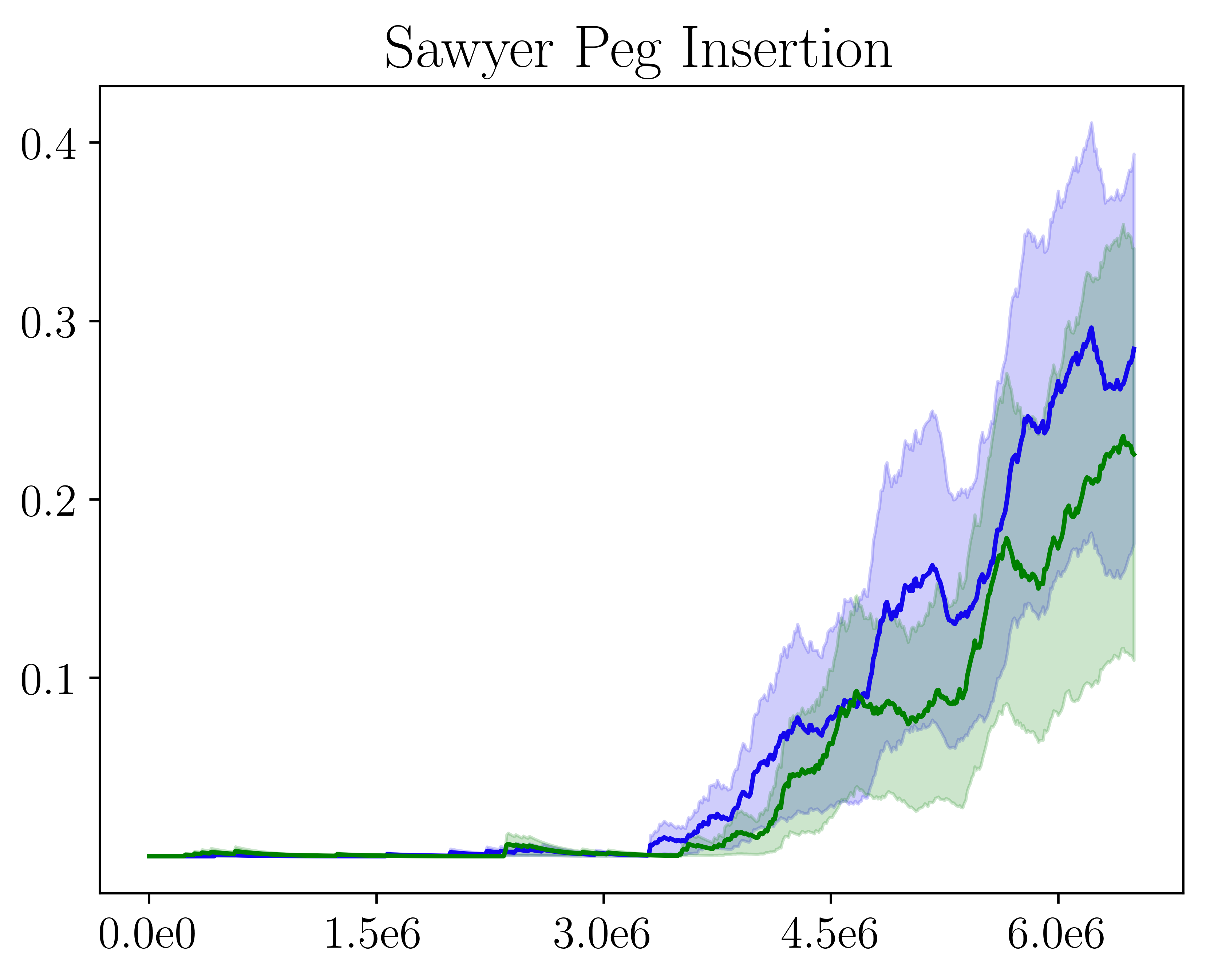}
    \includegraphics[width=0.4\textwidth]{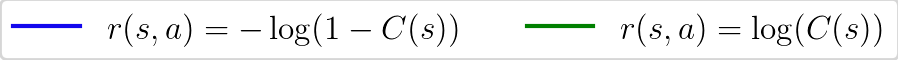}
    
    \caption{Performance comparison of $r(s, a) = \log(C(s))$ and  $r(s, a) = -\log(1-C(s))$ on (\textit{left}) the \textit{table-top organization} environment, (\textit{center}) the \textit{sawyer door closing} environment, and (\textit{right}) the \textit{sawyer peg} environment. Plots show learning curves with mean and standard error over 5 random seeds.}
    \label{fig:appendix_comparison}
\end{figure*}

Another choice that improved the stability was the choice of reward function for the backward controller: both $r(s, a) = -\log(1-C(s))$ and $r(s, a) = \log(C(s))$ preserve the saddle point $(\rho^*, 0.5)$ for the optimization in Equation~\ref{eq:minimax}. However, as can be seen in Figure~\ref{fig:appendix_comparison}, $r(s, a) = -\log(1-C(s))$ leads to both better and stable performance. We hypothesize that this is due to smaller gradients of the $-\log(1-C(s))$ when $C(s)\leq 0.5$, which is where the discriminator is expected to be for most of the training as the discriminator can easily distinguish between expert states and those of the backward policy to begin with.

\section{Environments}
\label{sec:append_environment}
The environment details can be found in \citep{sharma2021autonomous}. We briefly describe environments for completeness. For every environment, $H_T$ defines the number of steps after which the environment is reset to a state $s_0 \sim \rho_0$, and $H_E$ defines the evaluation horizon over which the return is computed for deployed policy evaluation: \\
\texttt{table-top organization}:
Table-top organization is run with a training horizon of $H_T = 200,000$ and $H_E = 200$. The sparse reward function is given by:
$$ r(s, g) = \mathbb{I}(\lVert s - g \rVert_2 \leq 0.2),$$
where $\mathbb{I}$ denotes the indicator function. The environment has $4$ possible goal locations for the mug, and goal location for the gripper is in the center. EARL provides a total of $12$ forward demonstrations and $12$ backward demonstrations ($3$ per goal).

\texttt{sawyer door closing}:
Sawyer door closing is run with a training horizon of $H_T = 200,000$ and an episode horizon of $H_E = 300$. The sparse reward function is:
$$ r(s, g) = \mathbb{I}(\lVert s - g \rVert_2 \leq 0.02),$$
where $\mathbb{I}$ again denotes the indicator function. The goal for the door and the robot arm is the closed door position. EARL provides $5$ forward demonstrations and $5$ backward demonstrations.

\texttt{sawyer peg}:
Sawyer peg is run with a training horizon of $H_T = 100,000$ and an episode horizon of $H_E = 200$. The sparse reward function is: 
$$ r(s, g) = \mathbb{I}(\lVert s - g \rVert_2 \leq 0.05),$$
where $\mathbb{I}$ again denotes the indicator function. The goal for the peg is to be placed in the goal slot. EARL provides $10$ forward demonstrations and $20$ backward demonstrations.

\end{document}